\documentclass{article}
\usepackage[a4paper, margin=1in]{geometry} 
\usepackage{graphicx} 
\usepackage{multicol}
\usepackage{array}
\usepackage{hyperref}
\usepackage{outlines}
\usepackage{authblk} 
\usepackage{footmisc} 

\newcolumntype{L}{>{\centering\arraybackslash}m{5.5cm}}

\usepackage{tcolorbox}

\usepackage[misc,geometry]{ifsym}
\usepackage{lineno}

\title{Evaluating General-Purpose AI with Psychometrics}

\author[1]{Xiting Wang}
\author[2,1]{Liming Jiang}
\author[3,4]{Jose Hernandez-Orallo}
\author[5,6]{David Stillwell}
\author[5,6,\Letter]{\\Luning Sun}
\author[2,\Letter]{Fang Luo}
\author[1,\Letter]{Xing Xie}

\affil[1]{Microsoft Research Asia}
\affil[2]{Beijing Normal University}
\affil[3]{Universitat Politècnica de València}
\affil[4]{ValgrAI}
\affil[5]{The Psychometrics Centre, University of Cambridge}
\affil[6]{Cambridge Judge Business School}
\date{}
\begin{document}

\maketitle

\footnotetext{\Letter$\:$Correspondence to Luning Sun (l.sun@jbs.cam.ac.uk), Fang Luo (luof@bnu.edu.cn), and Xing Xie (xing.xie@microsoft.com)}

\section*{Abstract}

Comprehensive and accurate evaluation of general-purpose AI systems such as large language models allows for effective mitigation of their risks and deepened understanding of their capabilities. Current evaluation methodology, mostly based on benchmarks of specific tasks, falls short of adequately assessing these versatile AI systems, as present techniques lack a scientific foundation for predicting their performance on unforeseen tasks and explaining their varying performance on specific task items or user inputs. Moreover, existing benchmarks of specific tasks raise growing concerns about their reliability and validity. To tackle these challenges, we suggest transitioning from task-oriented evaluation to construct-oriented evaluation. Psychometrics, the science of psychological measurement, provides a rigorous methodology for identifying and measuring the latent constructs that underlie performance across multiple tasks. 
We discuss its merits, warn against potential pitfalls, and propose a framework to put it into practice. Finally, we explore future opportunities of integrating psychometrics with the evaluation of general-purpose AI systems.\looseness=-1

\section{Introduction}


In the rapidly evolving field of artificial intelligence (AI), rigorous evaluation of AI systems allows for effective mitigation of risks associated with the AI systems and 
deepened understanding of 
their underlying 
technologies~\cite{Eisenstein2023}. It is crucial for preventing devastating outcomes in high-stake applications like autonomous driving and medical diagnosis~\cite{Burnell2023Rethink},
eliminating risks such as the propagation of racist, sexist, ableist, extremist, and other harmful ideologies~\cite{bender2021dangers}, and avoiding the reinforcement of biases against marginalized communities~\cite{dev2021gender,oliva2021fighting}. It also plays a key role in examining the extent and nature of intelligence AI systems possess~\cite{Eisenstein2023,hernandez2017measure}, preventing the misallocation of resources by reducing hype and misconceptions about AI capabilities~\cite{bender2021dangers}, and guiding improvements in future model iterations~\cite{binz2023using}. Recent policies such as the United States Executive Order 14110~\cite{uspolicy} underscore the importance of AI evaluation, highlighting its significance in ensuring responsible AI deployment.\looseness=-1

Developing a rigorous evaluation methodology is challenging, particularly considering the recent advancement in general-purpose AI systems. 
Unlike traditional AI systems tailored for specific tasks such as chess playing or code completion, general-purpose AI systems like 
ChatGPT~\cite{openaichatgpt} and Gemini~\cite{google2023gemini} demonstrate versatility in handling diverse tasks specified through text or multi-modal inputs such as audio, image, and video.
This versatility allows the systems to solve problems within complex and dynamic scenarios where tasks are previously unseen, rendering the mainstream task-oriented evaluation that assesses AI systems on predefined tasks less effective.
Take ChatGPT as an example: it can respond to novel tasks such as writing a mathematical proof in the style of a Shakespeare play~\cite{bubeck2023sparks}.
It is difficult to anticipate where users would apply the system and impossible to test every potential task to ensure that the system functions as intended. 
Consider also generalist medical AI models that may carry out unanticipated tasks set forth by a user for the first time~\cite{moor2023foundation}. Even if we have thoroughly tested the system on a number of  predefined tasks, we may still
misestimate when it fails 
due to its versality~\cite{moor2023foundation}.\looseness=-1

To tackle these challenges, we suggest transitioning from task-oriented evaluation to construct-oriented evaluation. Constructs are concepts that, although not observed directly, are hypothesized to underlie a range of behaviors~\cite{bacharach1989organizational,embretson2013item}. We follow the common assumption in behavioral science~\cite{bacharach1989organizational} that the wide range of outputs from an AI system can be organized and explained by a relatively small number of constructs, in the same way that human behaviors are explained by a reduced set of constructs such as cognitive abilities and personality traits~\cite{rust2020modern}\footnote{
Thurstone went further by stating that ``to deny this faith is to affirm the primary chaos of nature and the consequent futility of scientific effort.''~\cite{thurstone1947multiple}}. 
Based on the concepts of constructs, which are fundamental building blocks of hypotheses and theories~\cite{bacharach1989organizational}, we aim to advance the evaluation of general-purpose AI systems, enabling accurate prediction and explanation of their performance.\footnote{In the subsequent sections, the term ``AI''  by default refers to ``general-purpose AI,'' which is the primary focus of our paper, except where explicitly stated otherwise.
} 

To achieve this, constructs must be defined unambiguously and measured through empirical observations~\cite{wellington2007research}. 
In this paper, we draw expertise from psychometrics~\cite{rust2020modern}, which has studied human psychological constructs for over a century. 
Psychometrics is concerned with how constructs are empirically related to observable indicators such as test outcomes and involves the objective measurement of latent constructs. Unlike existing research that assumes the presence of human-like traits in AI systems and applies psychometric tests developed for humans as a tool to obtain ``psychological profiles'' of AI systems ~\cite{hagendorff2023machine,DOWE201277,hernandez2016computer,kosinski2023theory,pellert2023ai}, this paper is focused on leveraging the scientific theories and techniques in psychometrics to help identify and measure the constructs inherent in AI systems, be they human-like or not.\looseness=-1


This paper paves the way for a seamless integration of psychometrics with the evaluation of general-purpose AI systems. We first 
compare psychometrics with current task-oriented evaluation and detail its advantages, including predictive power, explanatory power, and quality assurance of the evaluation. We then examine the potential limitations of recent studies that employ psychometric tests developed for humans to evaluate AI systems. To guide construct-oriented evaluation, we present a framework grounded in psychometric principles and introduce how psychometric techniques can be incorporated and extended for AI evaluation. In the end, we discuss open questions and opportunities that warrant further investigation.
\section{The Advantages of Psychometrics in AI Evaluation}

\begin{figure}[t]
    \centering
    \vspace{-2mm}
    \includegraphics[width=1\textwidth]{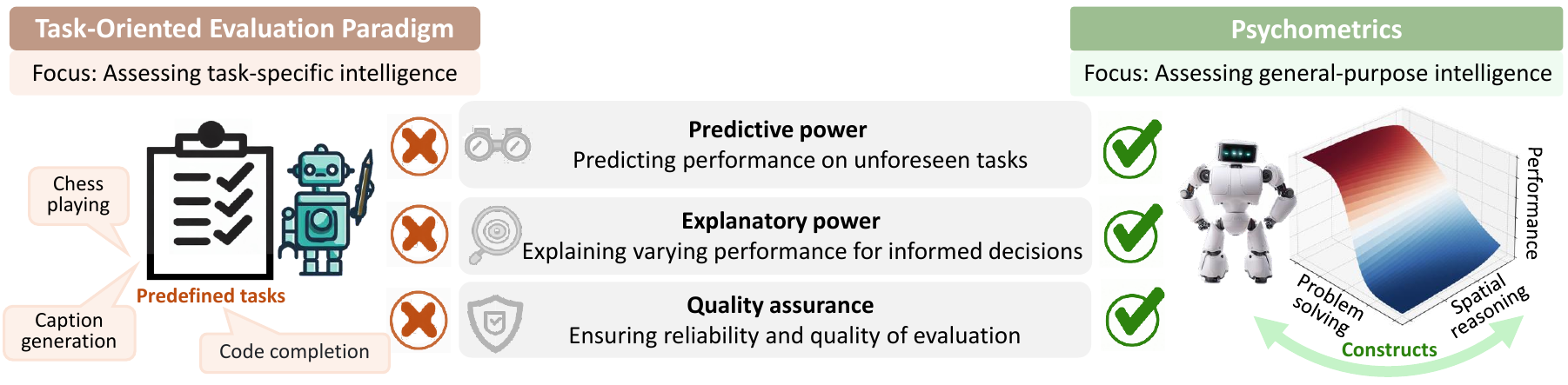}
    \caption{
    Comparison of the task-oriented paradigm for AI evaluation and psychometrics.
    }
    \label{fig:comparison}
\end{figure}

Given the versatility of current AI systems, big benchmarks that combine disparate tasks and tests have been created~\cite{srivastava2022beyond,liang2022holistic,gehrmann2022gemv2,chang2023survey,webb2023emergent}. For example, BIG-Bench~\cite{srivastava2022beyond} includes a collection of over 200 tasks. 
However, 
big benchmarks are unable to solve the problems intrinsic to task-oriented evaluation, which was originally designed to assess narrow, task-specific AI systems rather than general-purpose ones~\cite{hernandez2017measure}. 
As summarized in Figure~\ref{fig:comparison}, we present below the limitations of the task-oriented paradigm and explain how psychometrics, with a focus on latent constructs, provides a potential solution.

\subsection{Predictive Power}
\label{sec:issue}
Following the task-oriented paradigm, AI systems cannot be evaluated when the tasks are unforeseen or not pre-defined.  
For instance, although BIG-Bench contains tasks related to biology or medicine, it is insufficient to evaluate an AI system acting as a general medical assistant due to the variety of potential situations that could arise and the associated requirements. 
The assumption that AI's performance tested on a limited number of tasks 
directly reflect its performance in a practically infinite range of applicable tasks is unsubstantiated~\cite{hernandez2017measure,whiteson2011protecting}.

Psychometrics achieves predictive power by capitalizing on construct-oriented evaluation, as it has long been demonstrated that latent constructs are able to predict future performance and outcomes. For example, longitudinal human studies show that academic grades can be predicted from various latent constructs, such as cognitive ability, self-esteem, hope, and attributional style~\cite{Leeson2008Cognitive}. 
It is also reported that measures of personality predict a variety of consequential outcomes, including individual outcomes such as physical and mental health, interpersonal outcomes such as the quality of family relationships, and social institutional outcomes such as occupational choice, job satisfaction and criminal activity~\cite{ozer2006personality,Hoff2020Personality}. 
Based on latent constructs, psychometrics is capable of predicting human performance on a wide range of tasks and forecasting real-world outcomes.

Constructs of AI systems can be interpreted as descriptors for their performance across multiple tasks and scenarios. An example construct is spatial reasoning, which is required in a wide range of tasks involving planning and navigation, visual image interpretation, and natural language understanding~\cite{cohn2008qualitative}. 
If the goal is to create an algorithm that can safely drive a car, then it is reasonable to evaluate it on that narrow task. This is, after all, how we assess human driving ability. But if the goal is to predict whether an AI can solve several tasks that require spatial reasoning such as navigating a new environment, recognizing a rotated shape, and creating engineering blueprints for a new engine, then evaluating its spatial reasoning ability is more helpful.

Constructs that relate to multiple tasks offer a holistic understanding and maintain their relevance over extended periods~\cite{cronbach1955construct}. They allow the prediction of AI's performance even when the tasks are unforeseen. Imagine an AI system that proves excellent in constructs essential for a general medical assistant, for example, effective communication, clinical reasoning, and problem-solving ability. Without testing the system on specific tasks, we can confidently predict its performance as a general medical assistant, even in medical practices yet to be developed.

\subsection{Explanatory Power}
\label{sec:issue}

In order to identify the key strengths and weaknesses of AI systems, it is important to explain why they perform well or fail on certain tasks. However, current benchmarks are not intended to provide such an explanation, 
hindering researchers and policymakers from making informed decisions, for instance, to determine whether or when a system is safe and how it might be improved~\cite{Burnell2023Rethink}.

In contrast, psychometric tests are built upon latent constructs, which are conceptualized to explain empirical phenomenon~\cite{gopnik1994theory}. For example, according to the Cattell-Horn-Carroll theory, a large amount of individual difference in cognitive tasks can be explained by a relatively small number of cognitive abilities, such as fluid and crystallized intelligence~\cite{cattell1978check,carroll1997psychometrics}. These constructs form an underlying structure, which equips the psychometric test with explanatory power. 


Constructs are also useful to explain the performance of AI systems. For example, factor analysis reveals a consistent
structure in the performance of 29 different large language models on 27 cognitive tasks, which is comprised of three constructs representing reasoning, comprehension, and core
language modeling~\cite{burnell2023revealing}. These constructs explain a high
proportion of the variance in model performance and provide insights into the pattern of the capabilities across different models. 
Psychometrics provides rich techniques to identify explanatory constructs for AI systems (see Section~\ref{sec:construct-identification} for more details).

\subsection{Quality Assurance}

Serious concerns have been raised about the reliability and validity of task-oriented AI evaluation~\cite{mitchell2023how}. For instance, it has been frequently reported that the performance of AI systems is affected by factors such as input prompts and specific configurations~\cite{li2021prefix,jiang2020can,chen2023robust}.
The issue of sensitivity presents a challenge for benchmarks to serve as a reliable measure. It also questions the consistency of AI's performance in the real world. Moreover, it is hard to ascertain what is (not) being measured by a benchmark. 
For example, a high success rate of an AI system on a benchmark task may not translate into high performance in real-world applications. This disparity can be attributed to various factors, such as differences in task complexity between benchmarks and real-world applications, over-specialization in benchmarks, or biases in benchmark data sources~\cite{hernandez2017measure,whiteson2011protecting}.

Psychometrics has developed a systematic approach to test quality assurance, focusing on both reliability and validity of measurement. 
\emph{Reliability} demands stable and consistent results across multiple measurements~\cite{rust2020modern}. 
It goes beyond the scope of  ``replicability'' or ``robustness'' in computer science, offering a more comprehensive examination of different types of measurement errors, such as inconsistency of results among different test items, raters, and (sub-)datasets~\cite{xiao2023evaluating}. 
\emph{Validity} indicates the extent to which a test measures what it is designed to measure~\cite{whitely1983construct}. 
For example, a valid numerical reasoning test should tap into the ability of numerical reasoning rather than irrelevant constructs such as language proficiency. Only with reliable and valid tests can we put confidence in the measurement results and derive meaningful interpretations.

\section{Leveraging Psychometrics for AI Evaluation}

Thanks to the predictive power, explanatory power as well as rigorous quality assurance, construct-oriented evaluation that is guided by psychometric principles is becoming a necessary and promising paradigm for the evaluation of AI systems that are increasingly versatile. However, the integration of psychometrics with AI evaluation is not equivalent to simply applying existing psychometric tests to AI systems. Key considerations need to be addressed regarding how the constructs are conceptualized, how the tests are developed or adapted, and what psychometric techniques are available and appropriate. 
The ensuing discussion will elucidate both ineffective and effective strategies and raise open questions that warrant joint efforts from the wider academic community.

\subsection{The Perils of a Simplistic Application of Psychometrics}

Some recent literature has  
treated AI systems as participants in psychology experiments that were originally designed for humans~\cite{hagendorff2023machine,pellert2023ai} or applied 
existing psychometric tests to the evaluation of AI's general intelligence~\cite{DOWE201277,hernandez2016computer}, theory of mind~\cite{kosinski2023theory}, and personality~\cite{pellert2023ai}. 
The underlying assumption of these works is that AI systems such as large language models could exhibit human-like psychological traits, since they were trained on a large corpus that contains information about human values, beliefs, and personality traits~\cite{pellert2023ai}. 
However, this assumption might not hold because we do not know which human constructs are represented in AI systems, and neither do we know whether these constructs influence AI behavior in a manner akin to their impact on human behavior.
We discuss below reasons why employing tests developed for humans to evaluate AI systems could be misleading.


First, tests developed for humans may not be reliable or valid for AI systems, even if the systems are trained on large-scale human data. For example, several studies have applied self-report personality questionnaires that were originally developed for humans to large language models~\cite{li2022gpt}. 
Despite altered phrasing in the questions, as long as no actual semantic changes are introduced, human responses would remain consistent as they tend to reflect the personality traits. However, when applied to AI systems, a minor change in the input that is negligible to humans (such as reversing the order of the questions) may result in a substantial change in a large language model's response~\cite{li2022gpt}.    
This raises doubts about whether the responses of an AI system to self-report personality questionnaires reflect a true understanding of personality traits or are merely probabilistic selections based on the AI system's training data distribution.\looseness=-1

Second, it is problematic to assume that the relationship between the latent construct and its indicators identified in humans remains intact for AI systems. For instance, researchers consider processing speed an important indicator of intelligence in humans~\cite{drozdick2018wechsler}, as rapid thinking and response are often associated with high cognitive abilities~\cite{Embretson2010Measuring}. However, for AI systems, processing speed might not be relevant to intelligence or cognitive abilities. Smaller models that have fewer parameters typically process faster than larger models, but this does not necessarily mean that smaller models are more intelligent.

Therefore, when integrating psychometrics into the evaluation of AI systems, it is vital to reconsider the assumptions behind each psychometric test so as to determine its applicability~\cite{pellert2023ai}. In addition to adapting human tests, we would expect it to be necessary to develop new tests specifically tailored for the latent constructs inherent in AI systems, leveraging the principles of psychometrics.

\subsection{A More Rigorous Framework: Key Considerations}

\begin{figure}[t]
    \centering
    \includegraphics[width=1\textwidth]{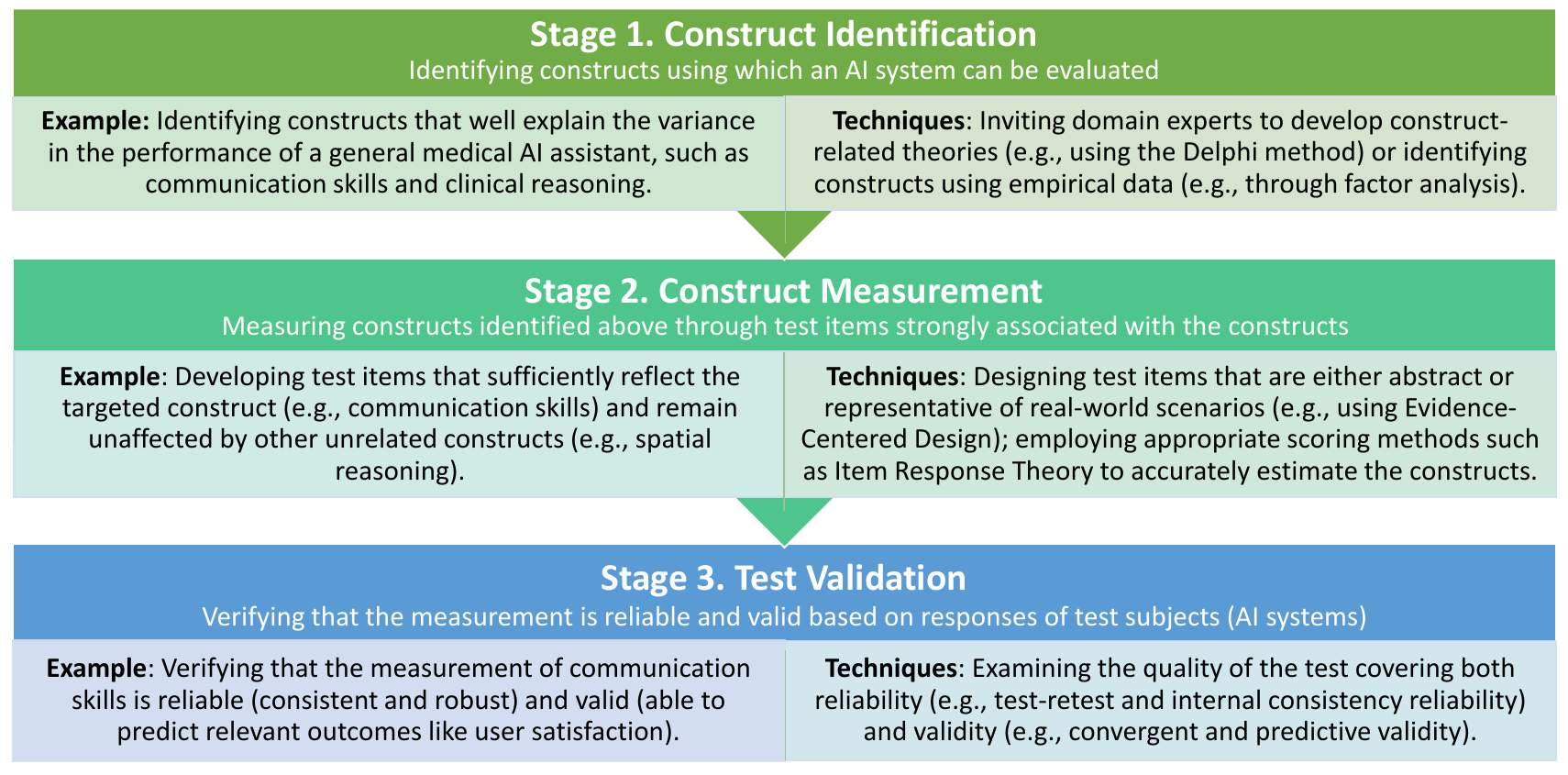}
    \caption{A framework for construct-oriented evaluation grounded in psychometrics, illustrated by an example of evaluating a general medical AI assistant, with exemplary psychometric techniques at each stage.\looseness=-1
    }
    \label{fig:framework}
\end{figure}

Here,  we suggest a framework for construct-oriented evaluation of AI systems. We introduce psychometric theories and techniques that could be employed to facilitate the evaluation. As shown in Figure~\ref{fig:framework}, our framework includes three stages: construct identification, construct measurement, and test validation.

\subsubsection{Construct Identification}
\label{sec:construct-identification}

In the first stage, we need to identify the constructs that can explain and predict AI performance. 
In psychometrics, there are essentially two types of approaches: top-down and bottom-up.

Top-down approaches start from a predefined theoretical framework, often derived from observations, expert consultation, and discussion~\cite{raykov2011introduction,masten1990resilience}. For instance, psychological resilience was identified through empirical observations that some children, despite facing adversities such as poverty or family discord, still demonstrated good adaptation and development~\cite{masten1990resilience}. 
Such insights are crucial in specifying the construct of interest~\cite{raykov2011introduction,masten1990resilience}. During the construct identification process, techniques such as the Delphi method~\cite{turoff2002delphi} can be adopted to encourage independent thinking, ensure anonymity, and circumvent group influence.

We can adopt similar approaches to identify constructs that are predictive and explanatory of AI's performance. 
For example, researchers have noted that AI systems often generate incorrect, fabricated, or misleading information~\cite{li2023halueval}, referred to as ``hallucination.'' In psychology, a similar concept is called confabulation, which might be leveraged as a construct to understand the above-mentioned behavior of AI systems~\cite{smith2023hallucination}. 
We can incorporate the observations with the expertise of domain experts to reach a consensus on the operationalized definition of confabulation in AI systems. 
The construct needs to be validated to ensure that it is predictive and explanatory~\cite{nunnally1994psychometric}. 

Drawing an analogy between humans and AI systems is one feasible top-down approach. However, we should notice that the definition and the importance of a construct may differ between AI systems and humans. 
Suppose that we are considering adapting the construct ``emotional intelligence'' for AI systems. We should adjust its definition to eliminate descriptions about self-awareness of emotions, which are crucial for humans~\cite{salovey1990emotional} but not AI systems. Instead, a definition of AI's emotional intelligence could focus primarily on the ability to recognize, understand, and respond to human emotions~\cite{wang2023emotional}.
The importance of a construct may also be different between humans and AI systems.
For example, while confabulation is prevalent among AI systems~\cite{smith2023hallucination}, confabulation may only manifest in a small group of individuals suffering from neurological conditions such as Alzheimer's disease or Wernicke-Korsakoff Syndrome~\cite{kopelman1997confabulation}.\looseness=-1 
Different from top-down approaches, bottom-up approaches start from empirical data and seek patterns that suggest the existence of a construct rather than relying on a predefined theoretical framework. Researchers may collect a wide range of behaviors or indicators and identify the underlying construct based on the relationships among these behaviors or indicators~\cite{raykov2011introduction}. A classic example is the development of the Big Five Factors of personality, where psychologists carried out factor analysis on a vast array of personality-descriptive terms 
and extracted five constructs that explain a large proportion of variance in human behaviors~\cite{goldberg1992development}. A similar approach has already been applied within the field of AI research~\cite{burnell2023revealing}. 
Notably, bottom-up and top-down approaches could be used in conjunction, complementing each other to provide a more comprehensive understanding of constructs. 

We should bear in mind that the conceptualization of constructs is constantly evolving. For instance, when the concept of self-esteem was initially introduced, it was primarily viewed as a single dimension, representing an individual's overall evaluation or feelings of self-worth~\cite{ea913da4-54ad-3e28-8ff8-bb2764065e0a}. As research progressed, psychologists began to recognize that self-esteem encompasses multiple facets and differentiated between state self-esteem (which relates to one's perception of self-worth in specific instances or situations) and trait self-esteem (which indicates a person's consistent or long-lasting level of self-worth)~\cite{8f1f4deb-3ffc-3f97-ad5e-391b5e83bc0b}. This evolvement in the understanding of constructs in psychology is also expected in AI evaluation. Newly identified constructs of AI systems are likely to be partial or biased. With enriched knowledge derived from empirical evidence, we should expect to continuously refine the definition of the constructs and their measurements. 


\subsubsection{Construct Measurement}
After identifying a construct, psychometricians design tests to measure it, which involves developing the test items and establishing the scoring criteria~\cite{groth2003handbook}. 

Test development is a challenging yet essential endeavor. Test items should not only sufficiently reflect the targeted construct but also remain unaffected by other unrelated constructs and factors~\cite{raykov2011introduction}. 
They are usually designed according to test guidelines~\cite{crocker1986introduction,haladyna2002review} and item specifications~\cite{popham1971criterion}. 
Test guidelines are the overarching framework for the development of a test. They specify the proportion of items measuring different aspects of the construct, the format of the items, the length of the test, time limits, and so on~\cite{crocker1986introduction}. 
Item specification pertains to the design of individual items. It specifies the source of item content, the context, stimuli, and the difficulty level of an item~\cite{crocker1986introduction}. 
Test guidelines and item specifications provide detailed design from the macro to the micro level, ensuring that test developers adhere to precise rules when crafting items. 

Test items for AI systems can take diverse formats as used in psychometrics~\cite{groth2003handbook}. 
For example, psychometric tests frequently use highly abstract symbols or graphics to eliminate the interference of unintended constructs introduced in real-world scenarios. When applied to AI evaluation, this could help minimize the unintended influence of random factors and enhance test reliability and validity~\cite{groth2003handbook}. 
For instance, to assess reasoning ability, we can develop abstract items like those in the number-series test~\cite{lefevre1986cognitive} that ask test-takers to uncover the pattern in a sequence of numbers and generate the number that continues the sequence. 
Another potential format is simulation-based~\cite{mislevy2013evidence}, which aligns more closely with current AI benchmarks.  
These test items employ real-world situations as context to increase ecological validity while leveraging techniques such as Evidence-Centered Design~\cite{mislevy2013evidence} to ensure that the target construct is measured effectively in such contexts. 
After the test items are developed, 
a scoring scheme based on appropriate theories~\cite{crocker1986introduction,baker2001basics} is established to quantify the level of the construct given the obtained responses.
Among existing scoring theories, Item Response Theory (IRT) appears particularly useful for AI evaluation. 
First, IRT effectively predicts whether a subject will get any item (including new unseen ones) correct.
This is achieved by modeling the probability of a correct response as a mathematical function of item parameters and the 
subject's level of the construct~\cite{embretson2013item, baker2001basics,hambleton1991fundamentals}. 
According to the function, if an item has a high difficulty parameter, then the 
subject will need to have a high level of the construct 
(e.g., higher reasoning ability estimated from previous answers) to get the item correct.  
IRT can be contrasted with traditional scoring methods such as Classical Test Theory (CTT)~\cite{crocker1986introduction}, in which correct items in the test are simply added up to come to a final score. This is the method used in most AI benchmarks, 
which does not model how a subject will perform on individual test items, especially unseen ones. It also suffers from the limitation that all items are weighted equally, despite that some items are evidently more difficult than others.
Secondly, IRT 
allows the construct and item parameters to be estimated on a unified scale~\cite{embretson2013item, baker2001basics,hambleton1991fundamentals}. 
This means that, for example, when assessing the logical reasoning of AI systems using IRT, the results of different AI systems each taking different sets of logical reasoning test items can still be compared directly. Given the rapid development of AI systems, frequent evaluation becomes imperative. Yet, using a fixed set of test items has been criticized due to the potential exposure of items in the training data~\cite{magar2022data}, so tests need to be constantly updated to avoid this issue.
IRT enables comparisons of AI systems even when they are evaluated in different tests with varying test formats or item sets~\cite{embretson2013item, baker2001basics,hambleton1991fundamentals}. 
Thirdly, IRT lays the groundwork for Computerized Adaptive Testing (CAT)~\cite{embretson2013item, baker2001basics}, which enables the selection of items that best match the construct level of each subject. 
This results in a more efficient and precise assessment of AI systems.\looseness=-1 

IRT models and CAT are receiving attention in AI evaluation~\cite{zhuang2023efficiently,martinez2019item}. We expect more insights to be gained from state-of-art IRT models, for example, IRT-based cognitive diagnostic models that aim at pinpointing specific facets of strengths or weaknesses in the construct~\cite{dibello2007guest} and IRT-based latent class models that combine the strengths of IRT with the idea of identifying the hidden groups that test takers belong to~\cite{roussos2007skills}. These techniques may result in a more fine-grained measurement of the targeted construct, allowing a more accurate and comprehensive evaluation of AI systems.

\subsubsection{Test Validation}

Imagine that we have created a test to measure a certain construct. Before drawing any conclusions based on the test results, it is necessary to ascertain the quality of the test. For instance, how much error is in the measurement? How do the test results relate to real-world behavior? For this purpose, psychometrics provides a systematic and rigorous methodology that focuses particularly on indicators of reliability and validity in the measurement.

\textbf{Reliability} refers to the consistency or stability of a test~\cite{rust2020modern, nunnally1994psychometric}. 
In psychometrics, there are multiple reliability indicators.
For instance, \textit{test-retest reliability} involves giving the same test to the same group of test takers multiple times to evaluate the stability of the measurements~\cite{rust2020modern, nunnally1994psychometric}. In practice, one could administer the test to an AI system multiple times to gauge its performance stability (i.e., replicability). 
\textit{Internal consistency reliability} is a measure of the consistency of results across items within a test~\cite{rust2020modern, nunnally1994psychometric}.  
High internal consistency reliability indicates that an AI system with a high ability on the latent construct should consistently exhibit good performance across all test items. 
Other types of reliability that might be adapted into AI evaluation include \emph{parallel forms reliability}~\cite{rust2020modern, nunnally1994psychometric}, which ensures two different but equivalent versions of a test measure the same construct consistently, and \emph{inter-rater reliability}~\cite{rust2020modern, nunnally1994psychometric}, which assesses the degree of agreement among different raters and has been adopted for evaluating text generation models~\cite{xiao2023evaluating}.

\textbf{Validity} indicates the extent to which a test measures the construct that it claims to measure~\cite{rust2020modern, nunnally1994psychometric}. There are different validity indicators in psychometrics. For example, \textit{construct validity} indicates how well the test reflects the latent construct. A commonly used method to evaluate construct validity is factor analysis~\cite{Gibson1959Three,Yong2013A}, which conceptualizes the constructs as latent factors underlying test items. By examining the fit of the specified structural model with empirical data, we can quantitatively understand the construct validity of the test~\cite{Gibson1959Three,Yong2013A}.
Construct validity can also be evaluated via two indicators: \textit{Convergent validity} reflects the extent to which two measures of constructs that are theoretically expected to be related are indeed empirically related; and conversely, \textit{discriminant validity} reflects if measurements that are supposed to be unrelated are, in fact, unrelated~\cite{nunnally1994psychometric,duckworth2011meta}. For instance, when evaluating the critical thinking of an AI system, we would expect its performance to be correlated with measures of its argumentation and not (or less) correlated with measures of unrelated constructs such as spatial reasoning.\looseness=-1

\textit{Predictive validity} is another form of validity that is concerned with the extent to which a score on a test predicts performance on a certain criterion measure~\cite{nunnally1994psychometric}. The criterion measure can be real-life performance or future outcomes. For example, individuals with higher intelligence are expected to achieve higher SAT scores~\cite{frey2004scholastic}. AI systems such as chatbots with stronger emotional understanding should receive better reviews from the users.
Both construct validity and predictive validity are important 
since they ensure that the construct being measured by the test is related to the actual use of the AI systems in real life. With adequate test validity and reliability, we would be able to trust the test results and interpret them to answer questions such as: ``Does a higher score in problem-solving tests for GPT-4 relative to GPT-3 mean a real increase of satisfaction among users when used to assist scientific research?'' 

\subsection{Open Questions}

Leveraging psychometric theories and techniques to evaluate AI systems brings forth both opportunities and challenges. Traditional psychometrics is designed for humans. 
Given evident differences between AI systems and humans, it is necessary to revisit some fundamental principles when applying psychometrics to the evaluation of AI systems.

\textbf{Redefining ``population'' and ``person'' in psychometrics}.
In psychometrics, within-person analysis is distinguished from between-person analysis, as dealing with multiple data points from one person involves totally different methods than dealing with a population where individuals contribute one data point each. 
For humans, the distinction between a person and a population is clear.
However, it is hard to define multiple data points collected from an AI system, which might be treated as either repeated measures or a single measure from multiple individuals. It is also noted that an AI system does not adhere to a single predefined role; instead, it has the capacity to take on an infinite variety of roles~\cite{shanahan2023role}. This makes the notion of  ``person'' and ``population''  
ambiguous for AI systems. For example, if we create different personas in the prompts~\cite{wang2023unleashing}, would this generate different persons? Is a fine-tuned AI model the ``same person'' as the previous model? If we consider different versions of models (e.g., GPT-3.5 and GPT-4) as ``multiple persons,'' do they belong to the same population?\looseness=-1

To determine whether data points generated by an AI system are a ``population'' or multiple data points from one ``person,'' we could examine the magnitude of the variance. When a test is taken by a population, the variance (i.e., between-person variance) is relatively large, as it reflects the diverse levels of the latent trait among individuals. In contrast, when the same person takes a test multiple times, the variance in the responses (i.e., within-person variance), which is caused by unstable factors such as mood, is considerably small~\cite{kenny1995trait}. 
Accordingly, we may consider an AI system under different settings as a single person generating repeated responses if the variance is small, and treat them as a population if the variance is sufficiently large. However, this raises the question of what is considered sufficiently large. One potential approach is to take the within-person and between-person variances observed in humans as a reference. Further research is required to determine the effectiveness of this approach and to explore other solutions.\looseness=-1 

Understanding the nature of the variance allows the employment of appropriate analysis methods.
For example, if multiple levels of data points are identified, we can adopt multilevel modeling to examine the effects at different levels~\cite{goldstein2011multilevel}, which may provide valuable insights into the performance variance of AI systems at the prompt level, persona level, and model level. 

\textbf{Handling prompt sensitivity}.
The pronounced sensitivity of AI systems to prompts~\cite{jiang2020can} highlights the relevance of prompt engineering and other elicitation techniques such as chain of thought~\cite{wei2022chain}. 
When evaluating an AI system, should we use the same instructions or prompts that are designed for humans, or should we make adjustments based on the characteristics of the AI system? Should the prompts represent how users will typically issue commands to an AI system or be carefully tuned to elicit optimal performance? If it is the latter, is it even possible to choose the most effective prompt from a large number of options? To address these questions, we need to broaden the scope of discussion and develop a standardized protocol that is accepted by the entire academic community.

On a related note, it remains a question how we should interpret the performance variations in AI systems caused by prompt engineering. 
Some prompts may pose a consistent impact on AI's performance. For example, prompts that prime a certain persona may result in a consistent gain in the performance of AI systems. 
Such a systematic effect could indicate that a second construct that we did not intend to measure may have been activated, affecting the validity of the measurement. 
Other prompts, for instance, created by paraphrasing, may lead to a random variation but have no effect on the average level of performance. This is a random effect that might affect the reliability of the measurement. 
Currently, it is uncertain what prompts would result in systematic or random effects, and systematic and random effects may also vary between different constructs and tests. In the future, further efforts are warranted to better understand and manipulate the effects of the prompts on AI systems.

The issue of prompt sensitivity also raises concerns regarding the stability of AI systems: Are these systems so unstable that they impede the development of a test that can evaluate them reliably? And more importantly, from a practical perspective, if the test items prove unreliable in evaluating an AI system, should we revise the test items to improve the reliability or optimize the AI system to mitigate the lack of reliability?
To address these questions, it requires collaborative efforts from researchers in both AI and psychometrics, and possibly a wider academic community. 

\textbf{AI vs. humans: A comparative exploration}.
In some cases, there is a need for tests that are applicable to both humans and AI systems, for instance, to benchmark AI systems with humans. To ensure a fair comparison, it is important to ascertain the appropriateness of the items. To address this challenge, we could learn from the technique of Differential Item Functioning (DIF)~\cite{osterlind2009differential} in psychometrics, which identifies instances where an item is biased against certain groups, after controlling for the level of the latent construct~\cite{osterlind2009differential}. DIF offers a method using which we could examine if a test item works equally well in humans and AI systems, as well as among different AI systems.

\section{Opportunities}


Building on the discussion in previous sections, we see further opportunities that psychometrics presents for AI 
research. Here we highlight how psychometrics might help shape the evaluation of human-AI teaming and transform the AI development pipeline in the future.




\subsection{Evaluation of Human-AI Teaming}

Given the demonstrated enhancements in human productivity facilitated by AI systems~\cite{noy2023Experimental}, it is increasingly evident that human-AI teaming may play an important role in the future of work. Initial research in this area has investigated Human-AI Decision Making~\cite{Lai2021Towards} and factors influencing the human-AI teaming process~\cite{Li2020Reciprocity,westerman2020thou}, while the evaluation of this hybrid form of work is still scarce. 

Teams of humans and AI systems, where humans are assisted by AI systems or collaborate with AI systems, could be regarded as a new form of working force. It is reasonable to adopt or adapt the principles mentioned in previous sections to guide the development of assessments for an accurate and efficient evaluation of their performance. Notably, human-AI teaming depends on the capacities of both the humans and the AI system to work in a complex and dynamic context~\cite{https://doi.org/10.1002/aaai.12108}, therefore, it is crucial to explore methods that disentangle the contributions of different types of agents in collaboration. 
In psychometrics, 
for example, researchers developed a collaborative problem-solving competency model to evaluate each participant's abilities, focusing on skills such as shared knowledge construction and negotiation. This model was validated through empirical studies in various contexts to assess performance in cooperative tasks~\cite{sun2020towards}. These methods could help with the evaluation of human-AI teaming.


 \subsection{Transforming the AI Pipeline}

\begin{figure}[b]
    \centering
    \includegraphics[width=1\textwidth]{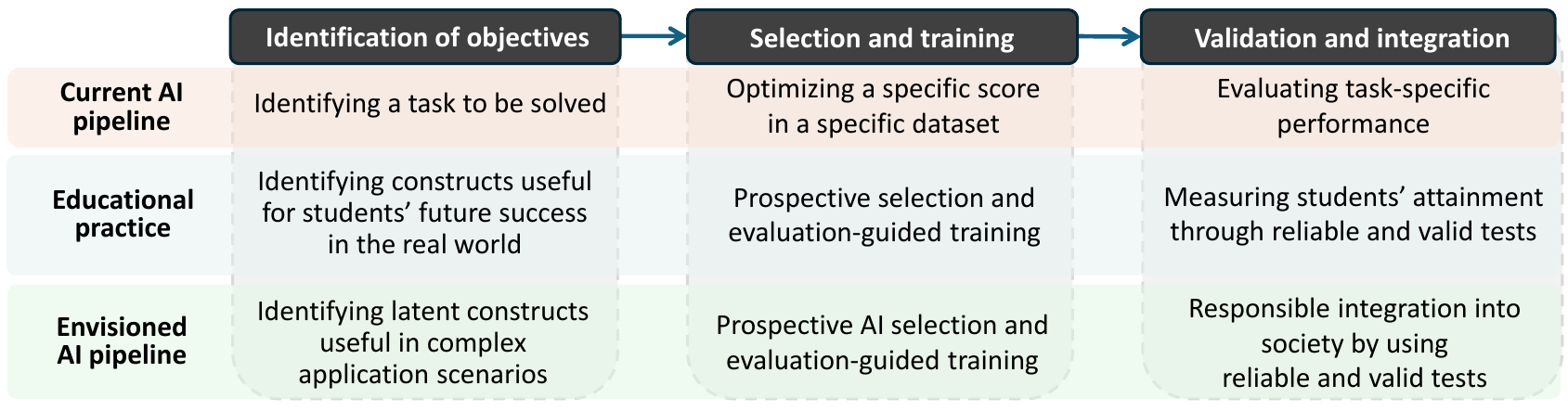}
    \caption{Comparison of the current AI pipeline, human education process, and a new AI pipeline supported by psychometrics. 
    }
    \label{fig:ai_pipeline}
\end{figure}

Assessment plays a critical role in informing and shaping human educational practices~\cite{kellaghan2001using}. For example, new teaching curricula are developed partly based on the assessment of current educational attainment. Similarly, the significance and role of AI evaluation should extend beyond measuring its performance after the system is trained. As required by the EU's Artificial Intelligence Act~\cite{europepolicy}, rigorous evaluation should guide each stage in AI development.   By drawing inspiration from how psychologists and educators integrate psychometrics with educational practice, we envision a future where psychometrics helps redefine the AI pipeline as shown in Figure~\ref{fig:ai_pipeline}.

When designing an educational scheme, educators start by identifying desirable objectives, which often include constructs that are in demand in real-world tasks~\cite{griffin2014assessment}. For example, the construct of problem-solving, i.e., the ability to solve complex and novel problems, is a 21st-century skill and considered a critical ability for students' future success in a rapidly changing world~\cite{trilling200921st}. Educational institutions, therefore, prioritize problem-solving as a key educational objective, which is then integrated into the teaching program as well as the assessment. For example, schools would incorporate problem-solving training within their teaching program, such as project-based learning that challenges students to apply their knowledge in a more practical and innovative manner~\cite{english201512}. Universities would also modify their entrance examinations to highlight problem-solving ability~\cite{fisher2005thinking}. The insights gained from the assessment would in turn inform the teaching strategies and help refine the curriculum to enhance students' problem-solving abilities based on identified weaknesses~\cite{popham1999classroom, wiggins1998educative}. Additionally, a rigorous examination with reliability and validity assurance would be conducted to confirm that a high enough level of problem-solving ability is indeed successfully obtained by students. 

Drawing inspiration from the educational practice, we may as well integrate construct-oriented evaluation in each step of the AI development. 
Specifically, before an AI system is developed, psychometrics can help identify the constructs required for use cases that are not limited to a single task or a combination of tasks like playing chess and classifying images.
In complex scenarios with unforeseen tasks, we may require AI systems to possess capabilities, such as critical thinking, and values, such as security and conformity as described in Schwartz's Theory of Basic Values~\cite{schwartz2012schwartz_theory}.
This holistic perspective based on constructs can ensure that AI systems align themselves with broader societal requirements in dynamic and complex scenarios.
In the selection and training stage, psychometrics can guide the selection of AI systems based on the insights derived from the evaluation. For example, psychometrics may help select AI systems that have the greatest potential for serving as a legal assistant.
If an AI system is capable of assessing the relevance of evidence to arguments and critically examining competing perspectives, there is a reasonable expectation that, even if it currently lacks extensive legal knowledge, it has the potential to assist a judge after being trained with substantial legal datasets. This prospective AI selection could save a large amount of training efforts and help allocate the most valuable resources to the most high-potential AI systems. 
Leveraging the feedback collected from the selection phase and ongoing tests, 
we can identify fundamental limitations 
(e.g., lack of critical thinking),
optimize the training towards the predefined objectives, and examine whether the AI system is improving over targeted latent constructs. 
In the validation and integration stage, psychometrics can facilitate the development of quality-assured tests for AI systems. 
By ensuring that the tests are both reliable and valid, we can verify that an AI system possesses the capabilities and values required for real-world challenges and responsible integration into society. 


\section{Conclusion}
When evaluating AI systems, current benchmarks, which are primarily task-oriented, fall short in key aspects such as predictive power, explanatory power, and quality assurance. 
This work contributes a perspective that 
addresses these limitations by placing psychometrics at the core of AI evaluation and focusing on latent constructs. 
Based on a three-stage framework, we have demonstrated how psychometrics can be incorporated, overcoming the limitations of current benchmarks while guarding against the potential pitfalls of oversimplified psychometric application. 
Integrating psychometrics not only promises significant advancements in AI evaluation, 
but also presents unique challenges, such as reinterpreting human-centric concepts for AI and managing prompt sensitivity. 
It opens up new avenues for research, including better evaluation of AI-human collaboration and the transformation of AI development practices. While we acknowledge that psychometrics alone may not solve all issues in AI evaluation, we consider this transition 
crucial in the current stage and look forward to a future in which psychometrics-grounded AI evaluation in return shapes the future of human evaluation.

\section*{Acknowledgments and Declarations}

\textbf{Acknowledgements}.
The authors would like to thank Jinyan Fan, Marija Slavkovik, Clemens Stachl, Alina A von Davier, Xiangen Hu, Yu Lu, Bryan Maddox, Mengxiao Zhu,  Meng Li, Li Dong, Jindong Wang, Igor Sterner, Greg Serapio-Garc\'{\i}a, Peter Romero, Fengli Xu, Fernando Mart\'{\i}nez Plumed, and Lidong Zhou for the discussions and comments on early versions of the manuscript.
This work was supported by the National Natural Science Foundation of China (Grant No. 62377003) and the Microsoft Research Asia Collaborative Research Program (FY23-Research-Sponsorship-422, ``The convergence of assessing human and big model capabilities'').
This work was also funded by the EU (FEDER) and Spanish grant RTI2018-094403-B-C32 funded by  MCIN/AEI/10.13039/501100011033, CIPROM/2022/6  funded by Generalitat Valenciana, EU's Horizon 2020 research and innovation program under grant agreement No. 952215 (TAILOR),
US DARPA HR00112120007 (RECoG-AI) 
and Spanish grant PID2021-122830OB-C42 (SFERA) funded by MCIN/AEI/10.13039/501100011033 and ``ERDF A way of making Europe". L.S. and D.S. gratefully acknowledge financial support from Invesco through their philanthropic donation to Cambridge Judge Business School.

\noindent \textbf{Author contributions}.
L.S., F.L., and X.X. jointly supervised this work. F.L. and X.X. initiated the project. X.W., J.H.O., D.S., L.S., F.L., and X.X. conceived the idea and formulated the structure. X.W. and L.J. drafted early versions of the manuscript. All authors were involved in the subsequent discussions and contributed significantly to the refinement of  the manuscript.

\noindent \textbf{Competing interests}.
The authors declare the following competing interests: At the time of writing, X.W. and X.X. were employees of Microsoft Research Asia and may own stock of Microsoft as part of the standard compensation package. Other authors declare no conflict of interest. 

\noindent \textbf{Additional information}.
Correspondence and requests for materials should be addressed to L.S., F.L., and X.X.
A preprint of this paper is available from \url{https://arxiv.org/abs/2310.16379}.
All the icons and images used in the paper are generated by \href{https://bing.com/create}{Image Creator in Bing}. 
\bibliographystyle{ieeetr}
\bibliography{main}

\end{document}